\documentclass[conference]{IEEEtran}
\IEEEoverridecommandlockouts
\usepackage{cite}
\usepackage{amsmath,amssymb,amsfonts}
\usepackage{algorithmic}
\usepackage{graphicx}
\usepackage{textcomp}
\usepackage{xcolor}
\usepackage{booktabs}           
\usepackage{multirow}           
\usepackage{amsfonts}           
\usepackage{graphicx}           
\usepackage{duckuments}         
\usepackage[colorlinks,linkcolor=blue,anchorcolor=blue,citecolor=blue]{hyperref}

\def\BibTeX{{\rm B\kern-.05em{\sc i\kern-.025em b}\kern-.08em
    T\kern-.1667em\lower.7ex\hbox{E}\kern-.125emX}}
\begin{document}

\title{InferTurbo: A Scalable System for Boosting Full-graph Inference of Graph Neural Network over Huge Graphs
}

\author{ 
	Dalong Zhang, Xianzheng Song, Zhiyang Hu, Yang Li, Miao Tao, Binbin Hu, Lin Wang, Zhiqiang Zhang, Jun Zhou$^{\ast}$ \thanks{*Corresponding author} \\
	\IEEEauthorblockA{\{dalong.zdl, xianzheng.sxz, zhiyang.hzhy, ly120983, taotao.tm, bin.hbb, fred.wl, lingyao.zzq,  jun.zhoujun\}@antfin.com} 
	\textit{Ant Group, Hangzhou, China}
}


\maketitle

\begin{abstract}
With the rapid development of Graph Neural Networks (GNNs), more and more studies focus on system design to improve training efficiency while ignoring the efficiency of GNN inference. 
Actually, GNN inference is a non-trivial task, especially in industrial scenarios with giant graphs, given three main challenges, i.e., scalability tailored for full-graph inference on huge graphs, inconsistency caused by stochastic acceleration strategies (e.g., sampling), and the serious redundant computation issue.
To address the above challenges, we propose a scalable system named InferTurbo to boost the GNN inference tasks in industrial scenarios.
Inspired by the philosophy of ``think-like-a-vertex", a GAS-like (Gather-Apply-Scatter) schema is proposed to describe the computation paradigm and data flow of GNN inference.
The computation of GNNs is expressed in an iteration manner, in which a vertex would gather messages via in-edges and update its state information by forwarding an associated layer of GNNs with those messages and then send the updated information to other vertexes via out-edges.
Following the schema, the proposed InferTurbo can be built with alternative backends (e.g., batch processing system or graph computing system).
Moreover, InferTurbo introduces several strategies like shadow-nodes and partial-gather to handle nodes with large degrees for better load balancing.
With InferTurbo, GNN inference can be hierarchically conducted over the full graph without sampling and redundant computation.
Experimental results demonstrate that our system is robust and efficient for inference tasks over graphs containing some hub nodes with many adjacent edges. 
Meanwhile, the system gains a remarkable performance compared with the traditional inference pipeline, and it can finish a GNN inference task over a graph with tens of billions of nodes and hundreds of billions of edges within 2 hours.

\end{abstract}

\begin{IEEEkeywords}
Graph Neural Networks, Distributed System, Full-graph Inference, Big Data 
\end{IEEEkeywords}

\section{Introduction}
Graph Neural Networks (GNNs) generalize deep learning over regular grids (e.g., images) to highly unstructured data and have emerged as a powerful learning tool for graph data.
Recently, GNNs have demonstrated impressive success on various tasks, ranging from traditional graph mining tasks (e.g., node classification\cite{b1,b2,b3,b4} and link prediction\cite{b5,b6}) to the fields of natural language processing\cite{b7} and computer vision\cite{b8}.
Meanwhile, graph data forms the basis of innumerable industrial systems owing to its widespread utilization in real-world applications (e,g., recommendation\cite{b9,b10,ab11,ab12,ab13}, marketing\cite{b11,ab15}, fraud detection\cite{b12, ab16}), further facilitating the flourishing of GNNs in industrial communities.

The key idea of current GNNs is taking advantage of information aggregation schema, which can effectively summarize multi-hop neighbors into representations via stacking multiple GNN layers. Essentially,  the high computation cost exponentially growing with the number of GNN layers poses the non-trivial challenges for \emph{GNN Training and Inference} in industrial scenarios with huge graphs, containing billions
of nodes and trillions of edges. With the aim of the possibility of performing the highly scalable and efficient GNN training in real industrial scenarios, several recent efforts have been made in industrial-purpose graph learning systems, including single-machine systems with the cooperation between GPUs and CPUs (i.e., DGL\cite{b13,b14} and PyG\cite{b15}), distributed systems based on localized convolutions (i.e.,  Aligraph\cite{b16}) and pre-processed information-complete sub-graphs (i.e., AGL\cite{b17}). Unfortunately, the main focus of these systems is training GNNs efficiently on graphs at scale, while few attention has been paid to the inference phase of GNNs~\footnote{Different from online inference whose goal is guaranteeing the low latency of real-time online serving~\cite{aab17,aab18}, the focus of GNN inference in our paper aims at efficiently performing full-graph inference in the offline environment, which is ubiquitous in financial applications.}, which is also the core of an industrial graph learning system.



In current graph learning systems\cite{b13,b14,b15,b16}, the inference phase is conducted by blindly imitating the pipeline of the training phase. However, the inference phase of GNN has its  unique characteristics, making it distinct from the training phase, such that the pure design of inference stage in current graph learning systems is unsuitable for performing large-scale inference tasks in industrial scenarios.
\begin{itemize}
    \item \emph{The large gap of data scale between training and inference phases.} 
    In a general graph learning task, only a small number of nodes (e.g., 1\% or even less in actual scenarios) are labeled for GNN training, while the inference tasks are usually expected to be taken over the entire graph. Such a gap is steadily worsening in the industrial scenarios with huge graphs, containing hundreds of millions or even billions of nodes\cite{b17, ab18, ab19} (e.g., Facebook\footnote{https://en.wikipedia.org/wiki/Facebook,\_Inc.}, Ant Group\footnote{https://en.wikipedia.org/wiki/Alipay}). Both observations encourage us to re-consider the scalability issue encountered in the inference tasks with the careful thought of memory storage and communication bandwidth.
    \item \emph{No guarantee for the consistency of predictions in inference phase.}
    As a basic requirement in industrial scenarios, the prediction score for a certain node should keep consistency at multiple runs, which unfortunately gets no guarantee in the GNN inference of current systems. To extend GNNs to large-scale graphs, the $k$-hop neighbor sampling strategy is widely adopted in current graph learning systems\cite{b13,b14,b15,b16,b17,b17append}. Although the training phase could greatly benefit from the efficiency-oriented strategy for generalized GNNs, the inherent stochasticity for prediction is unacceptable in inference phase,  especially for financial applications (e.g., fraud detection and loan default prediction\cite{b12, ab16}).
    \item \emph{Serious redundant computation issue in the inference phase.} 
    Current graph learning systems\cite{b13,b14,b15,b16} follow the procedure that performs forward and backward of GNNs over $k$-hop neighborhoods in a mini-batch manner, and would obtain a well-trained GNN by repeating this procedure enough times. In spite of its potential capability of enjoying mature optimization algorithms and data-parallelism in training phase, $k$-hop neighbor sampling with the mini-batch training manner would introduce the undesirable redundant computation issue in the inference phase since only one forward pass of GNNs is required. 
\end{itemize}

Addressing the limitations of current graph learning systems, we aim to facilitate GNN inference in industrial scenarios through a collaborative setting of mini-batch-based training and full-batch-based inference.
The significance of the cooperative setting is not trivial considering two major challenges that are not explored in current graph learning systems:
\begin{itemize}
    \item \emph{C1: How to unify the mini-batch based training and full-batch based inference in a graph learning system?} 
    As mentioned above, the mini-batch strategy is the promising way for efficient GNNs training, while the full-batch strategy is suitable for inference to alleviate computation issues. To combine those two advantages, it is impressive to abstract a new schema of graph learning that a GNN trained in the mini-batch manner could naturally perform inference in the full-batch manner.
    \item \emph{C2: How to efficiently perform GNN inference in distributed environments, especially for huge graphs with extremely skew degree distribution?}
    To ensure the consistency of prediction in inference tasks, we discard the strategies related to neighbor sampling, which put forward the urgent request for efficient GNN inference on huge graphs. In addition, due to the ubiquity of Power-Law, it is critically important to handle long-tailed nodes with extremely large degrees in terms of time cost and resource utilization, as well as avoiding the unexpected Out Of Memory (OOM) issues.
\end{itemize}

To this end, we design a system named \textbf{InferTurbo} to boost the inference tasks on industrial-scale graphs. 
We propose a GAS-like (Gather-Apply-Scatter) schema\cite{b19} together with an annotation technique to describe different stages of the data flow and the computation flow of GNNs, which could be used to unify the mini-batch training and full-graph inference phases.
The training and inference phases share the same computation flow but use different implementations of data flow.
The data flow parts are made as built-in functions currently, and users would only focus on how to write models in our computation flow.
Moreover, to handle the corner cases in natural graphs in industrial scenarios, such as nodes with extremely large amount of in-edges or out-edges, a set of strategies without any sampling like \emph{partial-gather}, \emph{broadcast}, and \emph{shadow-nodes} are proposed to balance the load of computation and communication and mitigate the long tail effect caused by those nodes.
As a result, consistency prediction results could be guaranteed at different runs, and the system enjoys a better load-balancing.
At last, the system is implemented on both the batch processing system\cite{b20,b21} and the graph process system\cite{b22}, which makes it gain good system properties (e.g., scalability, fault tolerance) of those mature infrastructures.
The main contributions of this work are summarized as follows:
\begin{itemize}
    \item We propose a GAS-like schema together with an annotation technique to describe the data and computing flow of GNNs, and make it possible to hierarchically conduct the inference phase over a full graph.
In this way, the redundant computation caused by k-hops neighborhood is eliminated in inference phase.
    \item We describe implementation details of the system on both batch processing system\cite{b20,b21} and graph computing system\cite{b22}, which are mature infrastructures and widely available in industrial communities.
    \item  We design a set of strategies such as \emph{partial-gather}, \emph{broadcast}, and \emph{shadow-nodes} for inference tasks to handle the power-law problem in industrial graphs.
    \item Compared with some state-of-the-art GNN systems (DGL, PyG), the proposed system demonstrates remarkable results in efficiency and scalability while achieving comparable prediction performance, and it could finish a node classification task over a graph with tens of billions of nodes and hundreds of billions of edges within 2 hours.  
\end{itemize}

%

\section{Preliminaries}
\label{sec:preliminaries}
\subsection{K-hop Neighborhood and Sampling}
A \emph{directed}, \emph{weighted}, and \emph{attributed} graph is denoted as
$\mathcal{G}=\{\mathcal{V},\mathcal{E},\mathbf{X},\mathbf{E}\}$, where
$\mathcal{V}$ and $\mathcal{E} \subseteq \mathcal{V} \times \mathcal{V}$ are the
node set and edge set of $\mathcal{G}$, respectively. 
$\mathbf{X}$ and $\mathbf{E}$ are node features and edge features.
The $k$-hop neighborhood $\mathcal{G}_v^k$ of node $v$, is defined as the \emph{induced attributed subgraph} 
of $\mathcal{G}$ whose node set is
$\mathcal{V}_v^k = \{v\} \cup \{u | d(v,u) \leq k\}$, where $d(v,u)$ denotes 
the length of the shortest path from $u$ to $v$
, and edge set is $\mathcal{E}_v^k = \{(u,u') | (u,u')\in \mathcal{E} 
\land u\in\mathcal{V}_v^k \land u'\in\mathcal{V}_v^k \}$. 
Additionally, $\mathcal{G}_v^k$ also contains the feature vectors of the nodes and edges. 
Without loss of generality, node $v$ itself is treated as its $0$-hop neighborhood.
It is proved that $k$-hop neighborhood would provide sufficient and necessary information for $k$-layer GNNs\cite{b17}.

However, the size of k-hop neighborhood grows exponentially with the number of hops, making the computations performed on it memory-intensive and time-consuming.
\emph{$K$-hop sampling} is proposed to address those problems. It would sample neighbors in a top-down manner, i.e., sample neighbors in the $k$-th hop given the nodes in the $(k-1)$-th hop recursively.
In particular, different sampling approaches\cite{b4, b10, b17, b17append, b17append2} employ different ways to select the neighbors, and one typical method is to randomly choose a fixed number of neighbors in each iteration.

\subsection{Graph Neural Networks}
Graph neural network is a popular way to learn over graph data, and it aggregates neighbors' information for a certain node in an iterative way.
Given an attributed graph $\mathcal{G}$, a GNN layer can be formulated in a \textbf{message passing paradigm}\cite{b13,b15,b23}, which first computes \emph{messages} via edges and then \emph{updates} the state of the target node with those messages:
\begin{equation} \label{eq:mpgnn}
\begin{array}{cl}
\text{message:} & m_{v,u}^{k+1} = \mathcal{M}(\mathbf{h}_v^k, \mathbf{h}_u^k, e_{v,u}),\\
\\
\text{update:} & \mathbf{h}_v^{k+1}= \mathcal{U}(\mathbf{h}_v^k, \mathcal{R}(\{m_{v,u}^{k+1}\}_{u \in \mathcal{N}^+_v})),\\
\end{array}
\end{equation}
where $\mathbf{h}_v^{k}$ denotes intermediate 
embedding of node $v$ at the $k$-th layer and $\mathbf{h}_v^{0} = \mathbf{x}_v$,
 $e_{v,u}$  and $m_{v,u}^{k+1}$ indicate edge features and messages associated with $v$'s in-edge from $u$ respectively,
$\mathcal{M}$ represents the \emph{message} function that generates messages according to adjacent node features and edge features on each edge,
$\mathcal{U}$ means the \emph{update} function that updates node embedding by aggregating incoming messages based on the \emph{reduce} function $\mathcal{R}$.
The \emph{message} function and \emph{update} function are usually neural networks, while the \emph{reduce} function could be a certain pooling function (such as sum, mean, max, min) or neural networks.

Most GNN algorithms (e.g., GCN\cite{b1}, GraphSAGE\cite{b4}, GAT\cite{b3}) could be expressed in this paradigm with a few changes in those functions.
For example, \emph{message} functions for GCN, GraphSAGE, and GAT, only take the source node's hidden state (or with edge features) as input.
In addition, \emph{reduce} functions for GCN and GraphSAGE are set to pooling functions, while for GAT, the reduce function would perform a weighted sum over messages based on the attention mechanism.

\section{Related Works}
\label{sec:relate_works}

\subsection{Graph Processing System}
The idea of message passing could be traced back to the graph processing field, which focuses on understanding the structure properties or topology of graphs, such as finding out the most important vertices (e.g., PageRank\cite{b24}), seeking the shortest path between nodes (e.g., SSSP\cite{b25}), and so on.
In the following, some gifted works will be introduced as they motivate our work a lot.

In the graph processing field, with the rapid growth of graph data, many efforts\cite{b26,b27} have been paid to conduct graph processing tasks distributedly to handle extremely large graphs.
Pregel\cite{b22} adopts a ``think like a vertex" programming model, in which each node receives messages from its neighbors in the previous iteration, modifies its own state or mutates the graph topology, and sends messages to other nodes via its out-edges.
PowerGraph\cite{b19} further develops the vertex-centric programming model with three fine-grained conceptual phases to describe the data flow, i.e., \textbf{G}ather, \textbf{A}pply, and \textbf{S}catter (\textbf{GAS}). 
The \emph{gather} and \emph{scatter} are a pair of symmetrical operations that demonstrate the information flowing in and out a node and perform edge-wise processing on the corresponding in-edges and out-edges.
The \emph{apply} stage is used to update the state of the node itself.
It enables a node-level paralleling by assigning nodes to partitions, along with their states and adjacent edges. Many graph processing algorithms (e.g., PageRank, SSSP) could be expressed in this paradigm.
The following works\cite{b28,b29,b30,b31} expand the abstraction and propose many talented strategies to optimize graph processing systems such as partitioning, communication, and so on.

The key difference between graph processing algorithm and GNN is that the former mainly focuses on graph structure, while the latter usually models graph structure and rich attribute information together and is optimized by gradient-based algorithms (e.g., SGD\cite{b32}).
That is, GNNs are more compute-intensive and communication-intensive, and data dependency exists in both forward and backward pass.

\subsection{Graph Learning System} 
\label{sec:gl_system}
Inspired by the development in deep learning communities, researchers also design and build systems to perform graph learning algorithms based on some deep learning frameworks, such as TensorFlow\cite{b33} and PyTorch\cite{b34}.
Many works\cite{b13,b14,b15,b16,b17} build graph learning systems following the message-passing paradigm to unify different GNN variants, which 
help researchers design new GNN algorithms in an easy way.  

The development of graph learning systems evolves from optimizing systems on a single machine\cite{b13,b14, b35,b36} to providing an efficient way to learn over large graphs distributedly\cite{b14,b15,b16,b17}.
Inspired by the k-hop sampling strategy\cite{b4,b10}, recent works\cite{b14,b15,b16,b17} perform localized convolutions on sampled k-hop neighborhoods to mitigate the inherent data dependency problem and enjoy the mature data-parallelism architecture in graph learning.
Specially, Aligraph\cite{b16} implements distributed in-memory graph storage engine, and workers will query subgraphs for a batch of training samples and do the training workloads.
DGL\cite{b14} proposes to use a partitioning algorithm when building the graph storage engine to reduce the number of edge cuts between different partitions, which is helpful to reduce the communication overhead when querying subgraphs of training samples.
PyG\cite{b15} provides lightweight, user-friendly sampling APIs, whereas users usually need to prepare an extra graph storage engine for distributed training.
AGL\cite{b17} proves that the k-hop neighborhood could provide sufficient and necessary information for k-layer GNN models and pre-generates the k-hop neighborhoods with a MapReduce pipeline to solve the inherent data dependency problem.
It also provides a way to estimate the errors caused by random sampling strategy, and errors would decrease as the sampling number increases.
Users might trade off computing speed and prediction accuracy by changing the number of neighbors for each layer.

However, little attention has been paid to the inference phase of GNNs.
Many works thought that inference tasks could be perfectly conducted in the same way as the training phase, which is hard to meet the requirements of inference tasks in industrial scenarios. The inference should be efficiently conducted over huge graphs with skewed degree distribution, and the prediction result should keep consistent at different runs.
A few existing works\cite{b37,b38} mainly focus on the efficiency problem in the inference phase on a single machine and propose some techniques, such as operation fusion and graph-specific caching, to reduce intermediate results and achieve better IO speed.
AGL\cite{b17} proposes a full-graph inference module to mitigate the redundant computation problem caused by k-hop neighborhoods but misses to solve the consistency problem and straggler problems caused by `hub' nodes in industrial scenarios.
It is still challenging to build an efficient and scalable inference system for industrial applications.

\section{System}
This section will first provide an overview of the InferTurbo inference system. The system is then detailed from three aspects: the programming model abstraction, implementations using various backends, and large-graph optimization techniques.

\subsection{System Overview}

The key motivation of the work is to build an efficient GNN inference system on top of scalability for huge graphs in industrial scenarios.
Therefore, instead of designing and optimizing inference system over a single monster machine, we'd rather build the system on mature, scalable infrastructures with high throughput capacity, such as batch processing systems and graph processing systems. 
In addition, to boost the inference efficiency, the system is expected to address the following issues: 
serious redundant computation caused by inferring on $k$-hop neighborhoods and the straggler problem brought on by skewed degree distribution.
Furthermore, as the consistency of prediction results at different runs is a fundamental requirement for industrial applications, some optimization strategies with randomness should be avoided.


\begin{figure}[htbp]
\centering
\includegraphics[width=\linewidth]{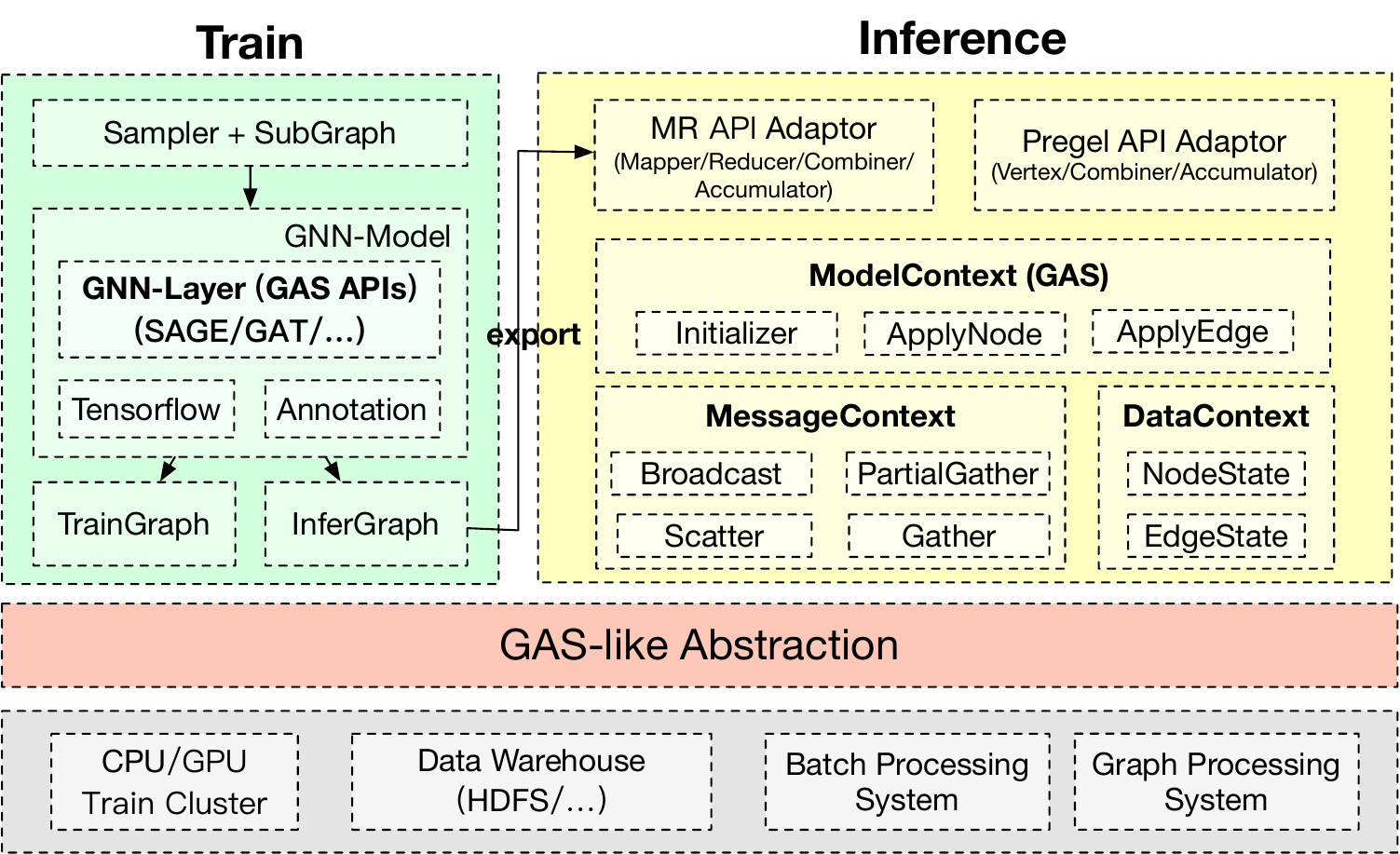}
\caption{System overview.}
\label{fig:system_overview}
\end{figure}


To this end, we propose InferTurbo to boost the inference tasks over industrial graphs.
The overall architecture is illustrated in Fig. 1.
First, a GAS-like abstraction is proposed to describe the data flow and computation flow of GNNs by combining the classical message-passing schema in \eqref{eq:mpgnn} and the GAS schema in graph processing system.
This abstraction is utilized to integrate the mini-batch training phase and full-batch inference phase.
In this way, the inference phase would not rely on k-hop neighborhoods and thus avoids redundant computation, while we could still enjoy the mature optimization algorithms in mini-batch manner and the data parallelism in training phase.



In addition, by designing and implementing adaptors, a well-trained GNN model in our abstraction could be deployed on batch processing systems or graph processing systems, which are mature industrial infrastructures with properties of high throughput and scalability.
Applications could choose one of them as the backend by trading off the efficiency and resource cost.

Furthermore, a set of strategies without dropping any information are proposed to handle problems caused by nodes with a large degree, since the degree distribution could be skewed for industrial graphs.
With those strategies, our system could achieve consistent prediction results at different runs and gain better load-balancing by mitigating the stragglers caused by those ``hub'' nodes.
\subsection{InferTurbo Abstraction}
In industrial scenarios, there are some key distinctions between the training and inference phases of GNN algorithms. 
In the training phase, the labeled nodes may be one percent of all nodes or even less, and the optimization procedure would be repeated several times on them to obtain a converged model.
It is a wise decision to conduct localized GNNs on those labeled nodes based on their k-hop neighborhoods since we can benefit from data parallelism and sophisticated optimization methods in the mini-batch setting.
The cost is also reasonable since labeled nodes would be scattered throughout the whole graph, and there would not be many overlaps between k-hop neighborhoods of corresponding labeled nodes.
On the contrary, inference tasks usually should be conducted on the entire graph.
Forwarding localized GNNs over the k-hop neighbors of all those nodes might result in the redundant computation problem.
A good way to solve it is to design a full-batch distributed inference pipeline and bridge the gap between those two different modes in training and inference phases. 

\begin{figure}
\centering
\includegraphics[width=\linewidth]{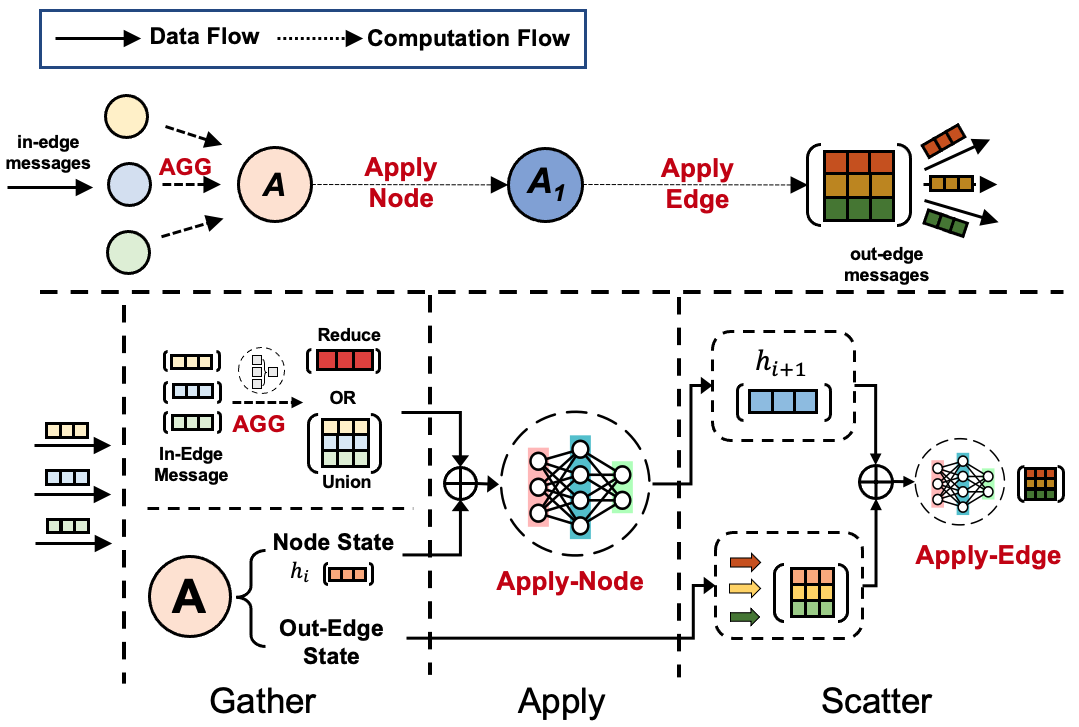}
\caption{InferTurbo abstraction}
\label{fig:gas_abstraction}
\end{figure}

\begin{figure*}
\centering
\includegraphics[width=.95\linewidth]{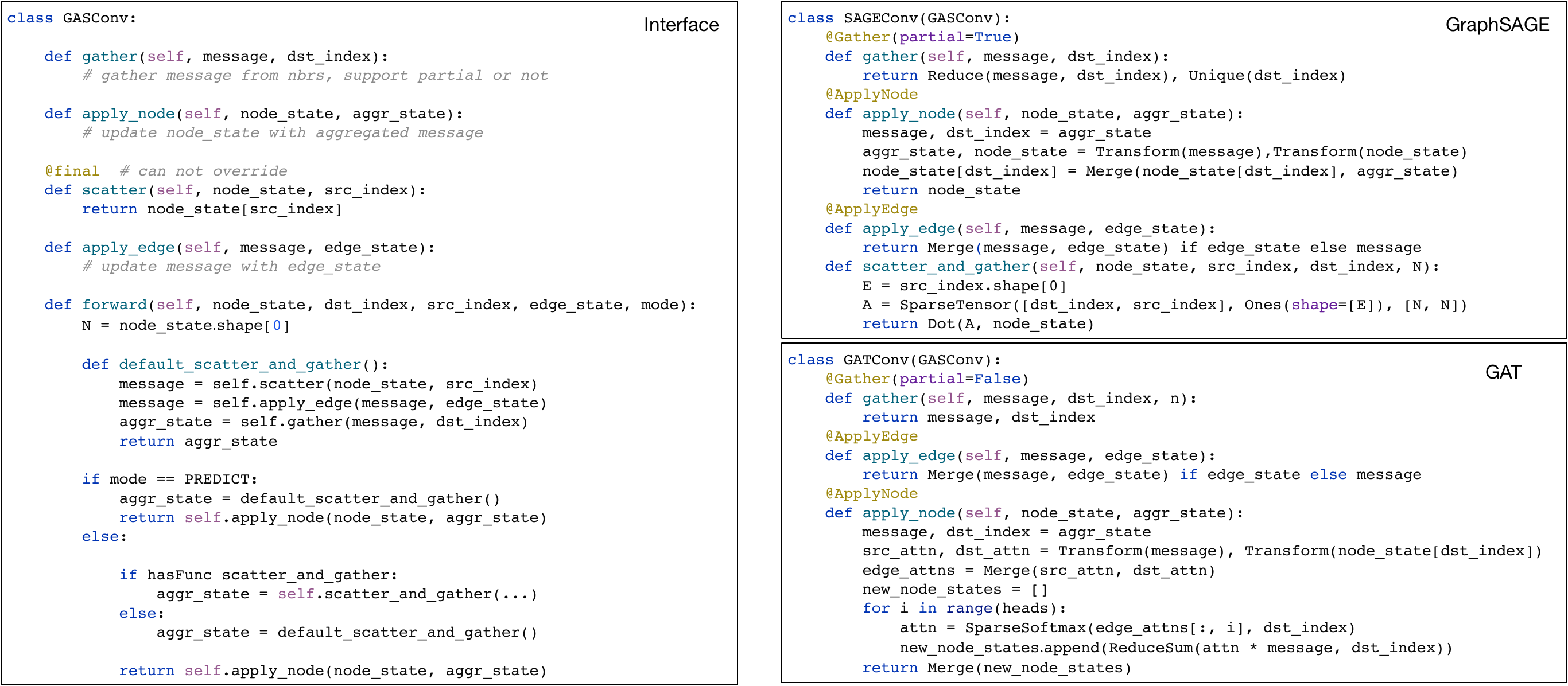}
\caption{GraphSAGE, GAT in InferTurbo abstraction}
\label{fig:gas_abstraction_example}
\end{figure*}



Inspired by the philosophy\cite{b22} of ``think-like-a-vertex" in the graph processing field, we re-express the classical message-passing schema of GNNs to a GAS-like abstraction to unify the mini-batch training and full-graph inference phases.
As shown in Fig.~\ref{fig:gas_abstraction}, for a certain GNN layer, the GAS-like abstraction can be defined as follows: 
\begin{itemize}
\item \textbf{Gather.}
	\begin{itemize}
	\item \emph{gather\_nbrs (\underline{Data Flow})}:
	A ``vertex" receives messages from its in-edges and then vectorizes the collected information into tensors. 
	\item \emph{aggregate (\underline{Computation Flow})}: 
	The ``vertex" would preliminarily process the messages from its in-edges, which is quite similar to the reduce function in $\mathcal{R}$ in \eqref{eq:mpgnn}.
	The difference is that this process should obey the commutative law and associative law (like max/min/sum/mean pooling or union) for further optimization in inference phase.
	Otherwise, the computation should be placed in the next stage.
	\end{itemize}
\item \textbf{Apply.}
\begin{itemize}
\item \emph{apply\_node (\underline{Computation Flow})}: The ``vertex" then updates its state information by combing the former state of itself and the ``gathered"  message from the former stage.
\end{itemize}
\item \textbf{Scatter.}
	\begin{itemize}
	\item \emph{apply\_edge (\underline{Computation Flow})}:
	The ``vertex" would generate messages according to the updated state information together with edge features.
	\item \emph{scatter\_nbrs (\underline{Data Flow})}:
	The ``vertex" sends messages via out-edges.
	\end{itemize}
\end{itemize}


Those five stages are used to describe the data flow and computation flow in GNNs, and their roles are emphasized by \underline{underlining} and annotating the corresponding part.
Compared with the classical GAS schema, both the \emph{Gather} and \emph{Scatter} are expanded to two sub-stages to distinguish the data flow and computation flow in those stages.
In general, \emph{gather\_nbrs} and \emph{scatter\_nbrs}, a pair of symmetry operations in the data flow, are used to receive and send messages via in-edges or out-edges, respectively.
We make them as built-in methods since they are the same among a variety of commonly used GNNs.

Computation stages would vary for different GNNs, and users should override those stages according to their specific requirements.
Specially, a rule is defined to set a boundary between \emph{aggregate} and \emph{apply\_node} stages: the computation of the \emph{aggregate} should obey the commutative law and associative law. Oherwise, related operations should be placed in the \emph{apply\_node} stage.
For example, many pooling functions (such as sum, mean, max, min), used as the reduce function in GCN and GraphSAGE, should be placed in the \emph{aggregate} stage following the rule.
For GAT, the computation of attention would break the rule, and thus, we simply union messages in the \emph{aggregate} stage and perform the reduce function in the \emph{apply\_node} stage.
This rule also facilitates further optimizations, which will be detailed in the next several sections.
The mini-batch training and full-graph inference are unified with such abstraction:
\subsubsection{Training}
In the training phase, following the traditional training-inference pipeline, the system still takes k-hop neighborhoods as input and train GNNs in the mini-batch manner.
As shown in Fig.~\ref{fig:gas_abstraction_example}, we take two widely used GNN algorithms (i.e., GraphSAGE, GAT) as examples to demonstrate how to organize codes in such abstraction.


The data flow in training phase is quite simple.
Since k-hops neighborhoods provide sufficient information for $k$-layer GNNs\cite{b17} and are locally available for a certain training worker, the data flow is just accessing and updating related local tensors.

Meanwhile, a model doesn't need much change and only should be expressed in our schema.
As shown in Fig.~\ref{fig:gas_abstraction_example}, a certain GNN algorithm should override three methods (\emph{gather}, \emph{apply\_node}, and \emph{apply\_edge}) from the base class.
In addition, we also provide an interface named \emph{scatter\_and\_gather} in case the \emph{scatter} and \emph{gather} stages could be fused together to avoid storing intermediate edge-level messages in training phase\cite{b13}.
For example, the scatter and gather processes in GraphSAGE are fused by a generalized sparse-dense matrix multiplication operation.
It's worth noting that since the \emph{gather\_nbrs} is just accessing local tensors in training phase, it is ignored here for simplicity.
The \emph{gather} interface in Fig.~\ref{fig:gas_abstraction_example} represents the computation of \emph{aggregate} stage.


Furthermore, some function decorators are developed to mark the beginning point and end point of functions, as shown in Fig.~\ref{fig:gas_abstraction_example}.
Meanwhile, we would generate layer-wise signature files to record those information at the time to save a well-trained model  (parameters and so on). 
In this way, different parts of the computation flow could be reorganized and deployed in corresponding stages in the inference phase.
Note that parameters in those decorators indicate whether to enable optimization strategies and also would be recorded in signature files.
Those information would be loaded in the inference phase to avoid excessive manual configurations.

\subsubsection{Inference}
Different from the training phase, the inference task is conducted in the full-graph manner.
Since the InferTurbo abstraction is expressed from the perspective of a node, the forward pass of GNNs could be treated as a ``vertex" program over a certain node.
By partitioning nodes in a graph into different machines, the total inference task could be conducted distributedly.

The data flow in the inference phase is quite different from that in the training phase.
Neighbors of a certain node could be placed on different machines, and data should be exchanged among those machines to prepare sufficient information to perform the computation flow on the node.
Therefore, rather than simply accessing local tensors, the data flow in inference phase mainly plays a role in communicating with other nodes on remote machines:
the \emph{gather\_nbrs} would receive information from remote nodes via in-edges and vectorize those information into adjacency matrix,  node/edge feature matrix, and so on.
The \emph{scatter\_nbrs} would send messages to other machines according to the destination node's id for the next iteration.

In contrast, the computation flow could be shared in training and inference phases.
In general, a certain computation stage of a well-trained model would be attached to the corresponding part in inference phase.
Specially, the computation flow would be reorganized for optimization.
For example, the \emph{aggregate} function may be performed in the \emph{Scatter} stage in advance to reduce messages sent to the same destination node.

The implementation details about how to conduct the inference tasks with specific backends and the optimization strategies will be presented in the following sections.

\subsection{Alternative Backends and Implementation Details}
Scalability is the key property that should be taken into consideration when implementing the InferTurbo in the inference phase.
Except for scalability, different industrial applications also have some specific requirements.
Inference tasks for some time-sensitive applications should be finished as fast as possible, even with relatively high costs on expensive and exclusive resources (like memory).
Others may be cost-sensitive and seek a way to conduct the inference over large graphs with limited commodity computation resources since the resource cost is also critical in industrial scenarios.

InferTurbo provides two alternative backends on graph processing systems (i.e., Pregel-like systems) and batch processing systems (i.e., MapReduce, Spark) to trade off computation efficiency and resource cost.
In general, InferTurbo on the graph processing system could be more efficient than that on batch processing system but with more strict requirements on stable and exclusive resources, such as, memory, CPU cores.
In contrast, InferTurbo on batch processing systems is more flexible in memory and CPU requirements as it processes data from external storage.
For example, the batch processing system could handle large graphs even with a single reducer if sufficient external storage is available.

The implement details of those two backends are introduced respectively as follows:

\begin{figure}
\centering
\includegraphics[width=0.9\linewidth]{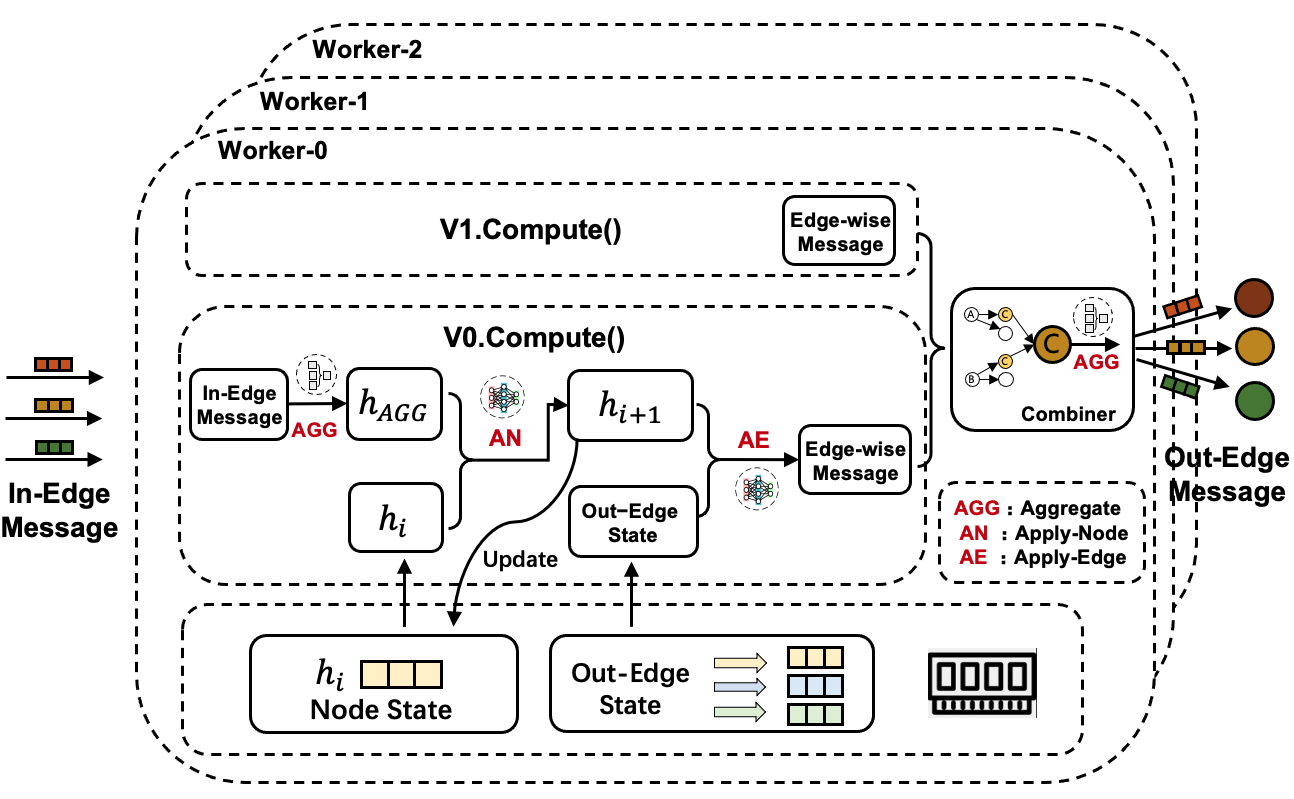}
\caption{The backend of a Pregel-like graph processing system}
\label{fig:backend_of_pregel}
\end{figure}

\subsubsection{InferTurbo on Graph Processing System}
We implement the InferTurbo abstraction on a Pregel-like graph processing system, which is one of the most popular distributed graph processing systems.
Note that it could be easily migrated to other graph processing systems as the abstraction originates from the graph processing field.



\textbf{Graph Partition.} Following Pregel\cite{b22}, a graph is divided into partitions according to node ids by a partitioning function (like, mod N), and each partition contains a set of nodes and all out-edges of those nodes.
Each partition is assigned to a worker to perform the ``vertex'' program (i.e., a layer of GNN).
As shown in Fig.~\ref{fig:backend_of_pregel}, a certain node would also maintain the node state and edge state (out-edges), such as raw features, intermediate embeddings, or even historical embeddings.
In this way, structure information and feature information could be stored in one place to avoid redundant storage.

It's worth mentioning that, with this partitioning strategy, the forward pass for a layer of GNNs over a certain node could be finished in one superstep by receiving messages from other nodes (maybe on other machines), which avoids synchronizing between different workers within a superstep.

\textbf{Execution.} 
As shown in Fig.~\ref{fig:backend_of_pregel}, we mainly perform the computation flow of GNN models in the \emph{compute()} function of the Pregel-like system. 

At first, we implement the \emph{gather} stage based on the \emph{message iterator} in Pregel, which is used to collect all messages sent to a certain node.
Meanwhile, we provide a built-in \emph{vectorization} function to transfer the messages and local state information into tensors, including structure information (i.e., adjacent matrix), node-level information (i.e., node state), and edge-level information (i.e., edge-wise messages or state).
Then, we perform the computation flow based on those matrices and provide functions to update node and edge states maintained in the machine.
At last, the messages generated by the model would be sent to other nodes via the \emph{send\_message} function in Pregel.
Specially, we design an optimization strategy for performing the aggregation part of the GNN model in the combiner stage of Pregel to reduce the communication cost, as shown in Fig.~\ref{fig:backend_of_pregel}.

After $k$ iterations (supersteps), we would finally get the result of a $k$-layer GNN model.
Note that the first and the last supersteps play special roles in the full pipeline.
The \emph{first} superstep could be treat as an initialization step since there are no messages received at this stage.
It mainly transforms the raw node states and edge features into initial embeddings and then calls the \emph{Scatter} stage to send those information to other nodes to start the following supersteps.
For the \emph{last} superstep, we attach a prediction part after the \emph{apply\_node} stage to get prediction scores for nodes (if needed) and then output the results (node embeddings or scores).

\begin{figure}
\centering
\includegraphics[width=\linewidth]{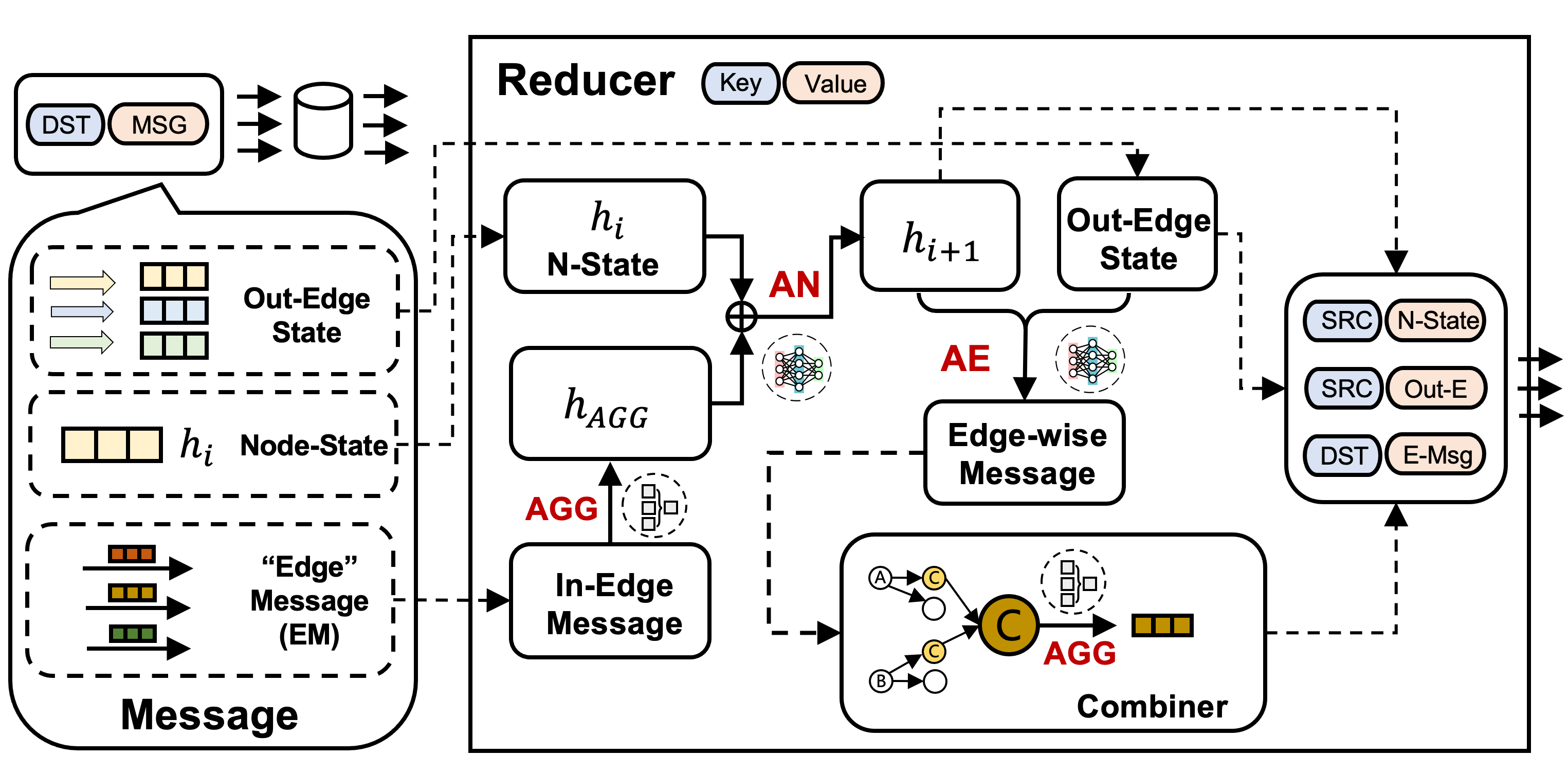}
\caption{The backend of MapReduce}
\label{fig:backend_of_mr}
\end{figure}

\subsubsection{InferTurbo on MapReduce}
We also provide an alternative backend on MapReduce (or Spark), as shown in Fig.~\ref{fig:backend_of_mr}.

Compared with the backend on the graph processing system, the MapReduce backend could not maintain some information (e.g., node states, out-edges) in memory.
Owing to this, the communication between different ``supersteps" could be quite different: all necessary information should be treated as messages and sent to the next round.

The MapReduce pipeline takes a \emph{node table} and an \emph{edge table} as input.
The \emph{node table}  consists of node id, node features, and ids of all out-edge neighbors, while the \emph{edge table} contains source node id, destination node id, and edge features.  
The overall pipeline on MapReduce is as follows:
\begin{itemize}
\item \emph{Map.}
The phase acts as the initialization step.
At first, it transforms the raw features of nodes and edges to initial embeddings.
Then, for a certain mapper, the initial embeddings of a certain node would be sent to itself and all its out-edge neighbors, while the edge information would be sent to both the source and the destination nodes.
In short, the \emph{Map} phase generates three kinds of information for each node: self-state information, in-edge information (i.e., edge features of in-edges, node features of the in-edge neighbors), and out-edge information (i.e., edge features of out-edges).

\item \emph{Reduce.}
The reduce phase performs the forward pass for a certain GNN layer.
Compared with the backend on graph processing system, the input and output in this phase are a little different.
In the \emph{Gather} stage, a node needs to receive edge messages and its own state. However, in the \emph{Scatter} stage, the node also sends the updated state as an additional message to itself.
That is, messages (e.g., node state, updated edge embeddings via out-edges) would be sent not only to destination nodes but also to itself.
The shuffle keys for those messages are set as ids of destination nodes or itself, respectively.
Messages for destination nodes represent in-edge information in the next round, while for itself, indicate new self-state information and out-edge information.
\end{itemize}

Similar to the implementation on Pregel, the prediction slice of a well-trained model will be merged to the last reduce phase, and after $k$ times \emph{Reduce} stages, we could finish the forward propagation of a $k$-layer GNN. 

Though the number of messages could be larger than that in the backends of graph processing system, 
this backend is promising for industrial applications since messages and state information usually are stored and exchanged with external storage (HHD or SSD).
As a result, a reducer could load a part of nodes together with their information into memory instead of loading all the data in the partition.
The peak usage of memory could be far less than that in the backends of graph processing system, as the latter backend usually should maintain all necessary information belonging to it in memory.
Therefore, it could be more suitable for scaling to extremely large graphs.
Moreover, many applications would leverage the ability of the powerful GNNs on various specific graphs in industrial scenarios.
The solution based on graph processing system is strict on stable and exclusive resources and may lead to resource competition for different applications.
The implementation on the batch processing system could release the heavy resource competition by sacrificing a little efficiency.

\begin{figure*}[htbp]
\centering
\includegraphics[width=\linewidth]{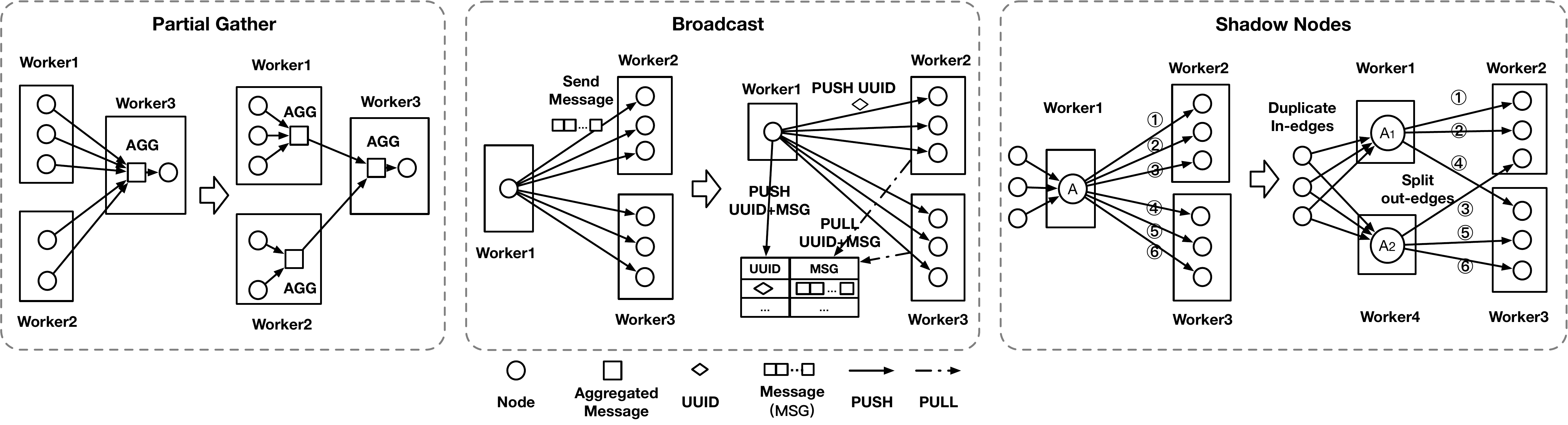}
\caption{Optimization strategies}
\label{fig:optimization}
\end{figure*}

\subsection{Optimization Strategies}
\label{sec_opt_stategies}
It is still challenging to conduct GNN inference tasks for industrial applications on those implementations since a hallmark property of graphs is \emph{skewed} power-law degree distribution\cite{b17,b19}.
The power law means a few nodes act as hubs and have many neighbors, while most nodes have few neighbors.
Those ``hub" nodes could increase the cost of computation and communication in inference phase, and substantially slow down the machines that contain them.
What's worse, the inference pipeline could crash as those nodes may lead to the OOM problem.
The neighbor-sampling strategy\cite{b4} may mitigate those problems but lead to unstable embeddings or prediction scores, which is unacceptable for industrial applications. 

By diving into those problems, we classify them into two categories: problems caused by the skewed in-degree and the skewed out-degree, and propose several strategies to address them respectively:

\paragraph{\textbf{Partial-Gather}}
For nodes with many in-edges, the time cost of receiving and computing over messages could increase significantly and lead workers containing them in the long tail.
We propose a \emph{partial-gather} strategy to address this problem, and the schematic diagram is illustrated in Fig.~\ref{fig:optimization}.

The basic idea of this strategy is to conduct the \emph{Gather} stage in advance to balance the computation of messages on the sender side while reducing the total amount of messages to be sent out. 
In addition, the five-stage GAS-like abstraction and the annotation technique described above make it possible to call 
related computation modules at any time in the pipeline.
Therefore, in a certain round, the \emph{aggregate} part of the model (i.e., gather function, partial=True) could be conducted using partially available messages when they are sent to the same destination.
Then in the next round, the receiver receives those aggregated messages from different workers and conducts the \emph{Gather} stage as usual.
Since the \emph{aggregate} function satisfies the commutative law and associative law, this strategy would not lead to any difference in final results.
We implement the strategy on both the Pregel-like system and the MapReduce (Spark) system with their built-in ``combining" function.

With this strategy, the communication complexity for a particular node is reduced to a constant level and only depends on the number of workers.
Additionally, the computation of \emph{Gather} is carried out relatively uniformly across source workers, which also avoids the computation hubs and may reduce the long-tail effect caused by large in-degree nodes.
This strategy involves almost no overhead and can be applied to all nodes regardless of their degrees.

\paragraph{\textbf{Broadcast}} 
For many GNN algorithms, all or part of messages sent via out-edges are usually the same. 
For example, the intermediate embeddings of a certain source node could be sent to all its neighbors via out-edges, which leads to overhead in communication.
For nodes with many out-edges, the communication would be the bottleneck, which should be avoided by ``compressing" those repeated messages.

To address this problem, we propose a \emph{broadcast} strategy as shown in Fig.~\ref{fig:optimization}.
In detail, instead of sending edge-level messages, nodes will send ``one" unique message to each machine together with a unique identifier (e.g., UUID or the source node id) as the index.
Meanwhile, identifiers would be sent as messages via out-edges.
Then, in the next round, the receiver should look up real messages by those identifiers and then perform the inference tasks as usual.
We implement this strategy with the built-in ``aggregator" class in batch processing systems and graph processing systems.
Furthermore, this strategy could be easily migrated on the latest graph processing systems which provide a native ``broadcast" mechanism.

It's obvious that the \emph{broadcast} could mitigate the overhead caused by the repeated messages and thus release the heavy communication pressure for nodes with large out-degree.

\paragraph{\textbf{Shadow Nodes}} 
A strategy named \emph{shadow nodes} is designed as an alternative solution for nodes with large out-degree, since the \emph{broadcast} strategy may not work well when messages vary with out-edges and cannot be compressed.

The basic idea of this strategy is to average the communication load caused by out-edges, as shown in Fig.~\ref{fig:optimization}.
First, the out-edges of a node are divided into $n$ groups evenly.
Then the node is duplicated $n$ times, and each mirror is associated with a group of out-edges and all in-edges. 
Note that the id of each mirror would be appended with the group information.
This process would be conducted in the preprocessing phase, and out-edges will be averaged to each mirror which could be placed on different machines.
Since each mirror holds all the in-edges of the original node and the out-edges of all the duplicates are equal to that of the original node, conducting GNN algorithms with this strategy is just the same as the original pipeline and would not change the result.
This strategy is simple but more general than the \emph{Broadcast} strategy.
It could re-balance the computation load whether messages could be compressed or not. 

Note that, in those two strategies for large out-degree problems, a heuristic formula is designed to estimate the threshold: $threshold = \lambda \times total\_edges/total\_workers$.
That is, those strategies should be activated when the number of out-edges for a node exceeds a certain percentage of edges on the worker.
In our experiment, the hyper-parameter $\lambda$ is set to an empirical value of 0.1.

\section{Experiment}
In this section, we conduct extensive experiments to evaluate our inference system and compare it with two state-of-the-art GNN systems, PyG\cite{b15} and DGL\cite{b13}.
\subsection{Experiment Settings}
\label{sec_exp_settings}
\textbf{Datasets.} 
Four datasets are used in the experiments, including PPI\cite{b39}, OGB-Products\cite{b40}, OGB-MAG240M\cite{b41}, and Power-Law, whose data scales range from small, medium, large, to extremely large.

The properties of those datasets are shown in Tab.~\ref{tab:datasets}.
The first three datasets are collected from the real world, and following the experimental settings in \cite{b3,b4,b17,b42}, they are divided into three parts as the training, validation, and test sets.
Note that settings of MAG240M follow the baseline examples provided by OGB team\footnote{https://ogb.stanford.edu/}, and only a part of the graph is used in the examples, and thus we count what we used in Tab.~\ref{tab:datasets}.

\begin{table}
\centering
	\caption{Summary of datasets}
	\label{tab:datasets}
	\begin{tabular}{c|cccc}
		\toprule
		Indices & PPI &Product & MAG240M & Power-Law \\ 
		\midrule
		\#Node   &56944  &2,449,029 &$1.2\times10^8$ & $1\times10^{10}$ \\
		\#Edge  &818716 &61,859,140 & $2.6\times10^9$ & $1\times10^{11}$ \\
		\#NodeFeature &50 &100 & 768 & 200 \\
		\#Class &121 &47 & 153 & 2 \\
		\bottomrule
	\end{tabular}
\end{table}

The Power-Law dataset is synthesized following the power-law\cite{b19} for the following two considerations.
Firstly, to verify the scalability of the system, graphs at different scales with the same degree distribution should be used in experiments, since the time cost could be affected a lot by different degree distributions even with the same scale.
Secondly, experiments for analyzing large in-degree and out-degree problems should be conducted on datasets with in-degree and out-degree following the power law respectively, for variable-controlling purposes. 
In our experiment, we synthesize datasets following the same rule with different scales and Tab.~\ref{tab:datasets} only records the largest one we used.
Specially, all nodes in Power-Law datasets are used in inference task, while millesimal are used in training phase.


\textbf{Evaluation.}
We design a series of experiments to evaluate our system in terms of effectiveness, efficiency, consistency, and scalability.

We first evaluate two widely used GNNs (GraphSAGE\cite{b4}, GAT\cite{b3}) on three real-world datasets to verify the effectiveness of our system.
Then, some experiments are conducted to show the unstable prediction scores caused by neighbor-sampling and prove the consistency of our inference system.
Moreover, we also record the time cost and compare it with traditional inference pipelines to see the efficiency of our system.


We verify the scalability of the system and analyze the effectiveness of optimization strategies on the Power-Law dataset.
We use \emph{cpu*min} to measure the usage of resources and record it at different data scales to prove the scalability of the system.
The data scales range from 100 million nodes (about 1 billion edges), 1 billion nodes (10 billion edges), to 10 billion nodes (100 billion edges). 
Specially, when analyzing the effectiveness of different optimization strategies individually,  power-law datasets are generated  with in-degree and out-degree following the power-law, respectively. 

The different backends of our system are deployed on different clusters. 
The backend on graph processing system consists of about 1000 instances, and each is powered by 2 CPU (Intel Xeon E5-2682 v4@2.50GHz) and 10GB memory, while the one on MapReduce contains about 5000 instances  (2-CPU, 2GB memory).
The network bandwidth is about 20 Gb/s.
For fairness, only 1000 instances are used in experiments to compare those two backends.


\subsection{Results and Analysis}
\label{sec_results_and_analysis}
\subsubsection{Comparisons}
We conduct a set of experiments to compare the system with traditional training-inference pipelines and present experimental results to demonstrate the effectiveness, efficiency, consistency, and scalability of the system.

\begin{table}
	\centering
	\caption{Performance}
	\label{tab:performance}
	\begin{tabular}{ccccc}
		\toprule
		 \multicolumn{2}{c}{Indices} & PyG &DGL &Ours  \\ 
		\midrule
		\multirow{3}{*}{SAGE} & PPI & 0.878 &0.878 & 0.880  \\
		\cmidrule(lr){2-2}
						&Product &0.787 &0.790 &0.788   \\
		\cmidrule(lr){2-2}
						& MAG240M & 0.662 &0.664 &0.668  \\
		\cmidrule(lr){1-5}				   
		\multirow{3}{*}{GAT}  & PPI & 0.987 &0.981 & 0.986  \\
		\cmidrule(lr){2-2}
						   &Product &0.794 &0.800 &0.801  \\
		\cmidrule(lr){2-2}
						   & MAG240M & 0.663 &0.659 & 0.670  \\
		\bottomrule
	\end{tabular}
\end{table}

\textbf{Effectiveness.} 
Effectiveness is always in the first place. Table~\ref{tab:performance} illustrates comparisons of performance between our system and traditional training-inference pipelines powered by PyG and DGL.
We report the results of GraphSAGE and GAT on PPI, OGB-Product, and OGB-MAG240M in different inference pipelines.
Configurations of those GNN algorithms follow examples in OGB leaderboard\footnote{https://ogb.stanford.edu/docs/lsc/leaderboards/}.

In general, results achieved by our system are comparable with those on PyG and DGL across different datasets and algorithms.
It's reasonable since our system just changes the way to perform the inference but never changes the formula of GNNs or introduces any approximation strategy in inference phase.
As those two algorithms represent the two most widely used algorithms while the scale of datasets ranges from small, medium, and large, those comparisons and results prove the effectiveness of our system in different scenarios.

%


\textbf{Efficiency.}
We evaluate the efficiency of our system by comparing it with the traditional inference pipelines powered by PyG and DGL, and record the resource cost measured by \emph{cpu*min} together with time cost on the OGB-MAG240M dataset.
Note that, in traditional pipelines, we use a distributed graph store (20 workers) to maintain the graph data and 200 workers (10-CPU, 10G memory) for inference tasks.
Increasing the number of workers would exacerbate the communication problem between workers and the graph store, reducing the efficiency of traditional pipelines.
For fairness, the total CPU cores of inference workers are equal to our system with different backends, and the CPU utilization remains 90\%+ during the inference tasks.


%

Table~\ref{tab:efficiency} demonstrates the experimental results compared with traditional inference pipelines.
Generally,  both the two backends of our system demonstrate superior efficiency.
Our system achieves 30$\times$ $\sim$ 50$\times$ speedup compared with traditional inference pipelines, while in terms of the total resource cost, traditional inference pipelines take about 40$\times$ $\sim$ 50$\times$ more than our system.
Those results show that our system is quite promising for industrial scenarios: it not only could finish inference tasks faster but also is more economy friendly for industrial applications as it takes fewer resources.
The results we achieved mainly benefit from the new inference pattern in our system.
That is, with the InferTurbo abstraction, it could conduct layer-wise inference over the entire graph while enjoying the good property of parallelism.
In this way, redundant computation caused by $k$-hop neighborhoods could be avoided.
Notably, even ignoring the bottleneck of communication in traditional pipelines and assuming they could achieve $5\times$ speedup by scaling up $5\times$ resource (i.e., 1000 workers), our system still gains up to $10\times$ speedup over them.

In addition, we also design a set of experiments to demonstrate the relationship between time cost and the number of GNN layers for further analysis, and results are presented in Tab.~\ref{tab:time_cost_hops}. 
Without loss of generality, the comparisons are mainly conducted between PyG (traditional pipeline) and the MapReduce backend.
\emph{nbr50} and \emph{nbr10000} mean the number of neighbors for a certain node is limited to 50 and 10000 respectively with neighbor-sampling strategy for experiments on PyG, while there is no sampling for our system.
Both the time cost and resource usage of our system increase nearly linearly by varying hops of GNNs, while they increase exponentially in the traditional pipeline.
Specially, the traditional pipeline even crashes with the OOM problem when the number of neighbors is set to 10000.
Because, in the forward pass of the traditional pipeline, the k-hop neighborhoods of nodes increase exponentially with the hop count, resulting in the exponential growth of communication and computation.
However, k-hops neighborhoods for different target nodes could overlap with each other, and the communication and computation on the overlaps are unnecessary.
Our system avoids those redundant problem by conducting inference tasks over the entire graph in the full-batch manner.
For each node on each layer, it will be used only once in our system.
Therefore, the time cost and resources are only related to the hop count.

\begin{table}
	\centering
	\caption {time cost and resource usage on different systems}
	\label{tab:efficiency}
		\begin{tabular}{cccccc}
		\toprule
		 \multicolumn{2}{c}{Indices} & PyG &DGL & On-MR & On-Pregel \\ 
		\midrule
		Time & SAGE & 780 &630 & 20 & 15 \\
		\cmidrule(lr){2-2}
		(min)				   & GAT & 1056 &948 & 34 & 21 \\
		\cmidrule(lr){1-6}				   
		Resource  & SAGE & $1.6 \times10^6$ &$1.3 \times 10^6$& $2.6 \times 10^4$ &$2.9 \times 10^4$  \\
		\cmidrule(lr){2-2}
		(cpu*min)				   & GAT &$2.1 \times 10^6$ &$1.9 \times 10^6$ &$4.4 \times 10^4$ &$4.1 \times 10^4$ \\
		\bottomrule
	\end{tabular}
\end{table}


\begin{table}
	\centering
	\caption{Time and resource cost VS. hops}
	\label{tab:time_cost_hops}
		\begin{tabular}{ccccccc}
		\toprule
		{} &\multicolumn{3}{c}{Time (min)} & \multicolumn{3}{c}{Resource (cpu*min)} \\
		\cmidrule(lr){2-4}
		\cmidrule(lr){5-7}
		hops& 1 &2 &3 & 1 &2 &3 \\
		\midrule
		 nbr50& 23 & 160 &3300+ &$4.5\times 10^4$ & $3.2\times 10^5$ & $6.7 \times 10^6$\\
		 nbr10000& 181 & 780 &{OOM} &$3.6\times 10^5$ & $1.6\times 10^6$ & {OOM} \\
		 \cmidrule(lr){1-7}	
		 ours &13 &20 &31 &$1.7\times 10^4$ &$2.6\times10^4$ &$4.0\times10^4$ \\
		\bottomrule
	\end{tabular}
\end{table}


\textbf{Consistency.} 
We take the GraphSAGE model in PyG on the OGB-MAG240M dataset as baselines to verify the consistency of our system and traditional pipelines.
The number of neighbors for a node in traditional pipelines is limited to 10, 50, 100, 1000 in different experiments.
We count the number of node classes predicted in the traditional pipeline at 10 runs and summarize such information for about 130,000 nodes to see whether predicting scores would keep unchanged.

\begin{figure}
\centering
\includegraphics[width=\linewidth]{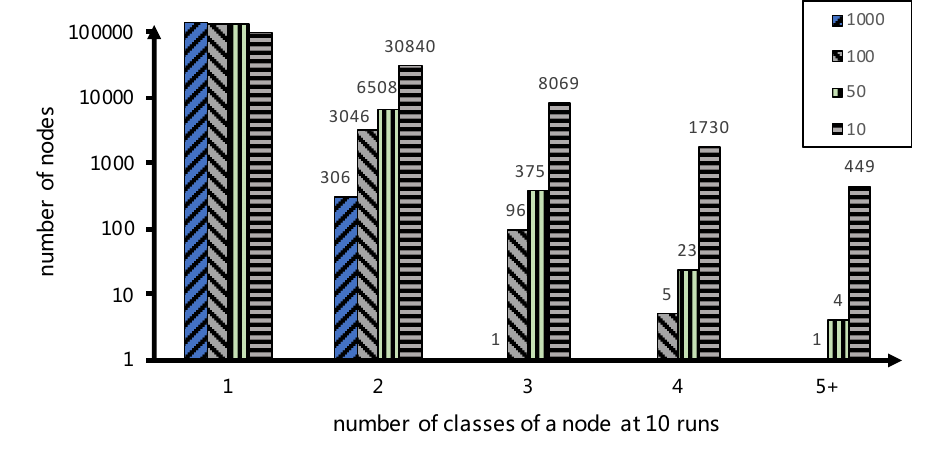}
\caption{Statistics of total number of classes a node predicted at 10 runs}
\label{fig:consistency}
\end{figure}

The statistical results are presented in Fig.~\ref{fig:consistency}.
The x-axis denotes how many classes a certain node belongs to, while the y-axis indicates how many nodes would be predicted to a certain number of classes at 10 runs.
For example, when the number of neighbors is set to 10, 30840 nodes are predicted into 2 different classes at 10 runs. 
In all, about 30\% nodes would be predicted to at least 2 different classes when the sampling number is set to 10, which indicates that the prediction score is not trusted.
Even increasing the sampling number to 1000, there are still about 0.1\% nodes in trouble, which is unacceptable in industrial scenarios, especially in financial applications.
Note that when the sampling number increases to 10000, such risk is mitigated, but the time cost increases a lot and even causes the OOM problem as shown in Tab.~\ref{tab:time_cost_hops}.

In contrast, our system achieves the same results at different runs since it conducts the inference phase over the full graph without sampling.
And we propose a series of strategies to avoid the OOM problem caused by nodes with a large degree.
Therefore, the system is suitable for inference tasks on large graphs, even with skewed degree distribution.

\textbf{Scalability.} 
Scalability is a key issue for industrial scenarios.
We design a series of experiments and illustrate the relationship between the resource usage (cpu*min) and the data scale to evaluate the scalability of our system.
The experiments are conducted over Power-Law datasets, and the data scale includes three orders of magnitude as mentioned in Sec.~\ref{sec_exp_settings}.
A 2-layer GAT model with an embedding size of 64 is conducted on those datasets, and we record the time and resource cost to see how they vary with different data scales.
Note that the backend of our system is set to MapReduce as the resource on the cluster for graph processing backend is not enough for the largest dataset.

\begin{figure*}[htbp]
\centering
\begin{minipage}{0.3\linewidth}
\centering
\includegraphics[width=\textwidth]{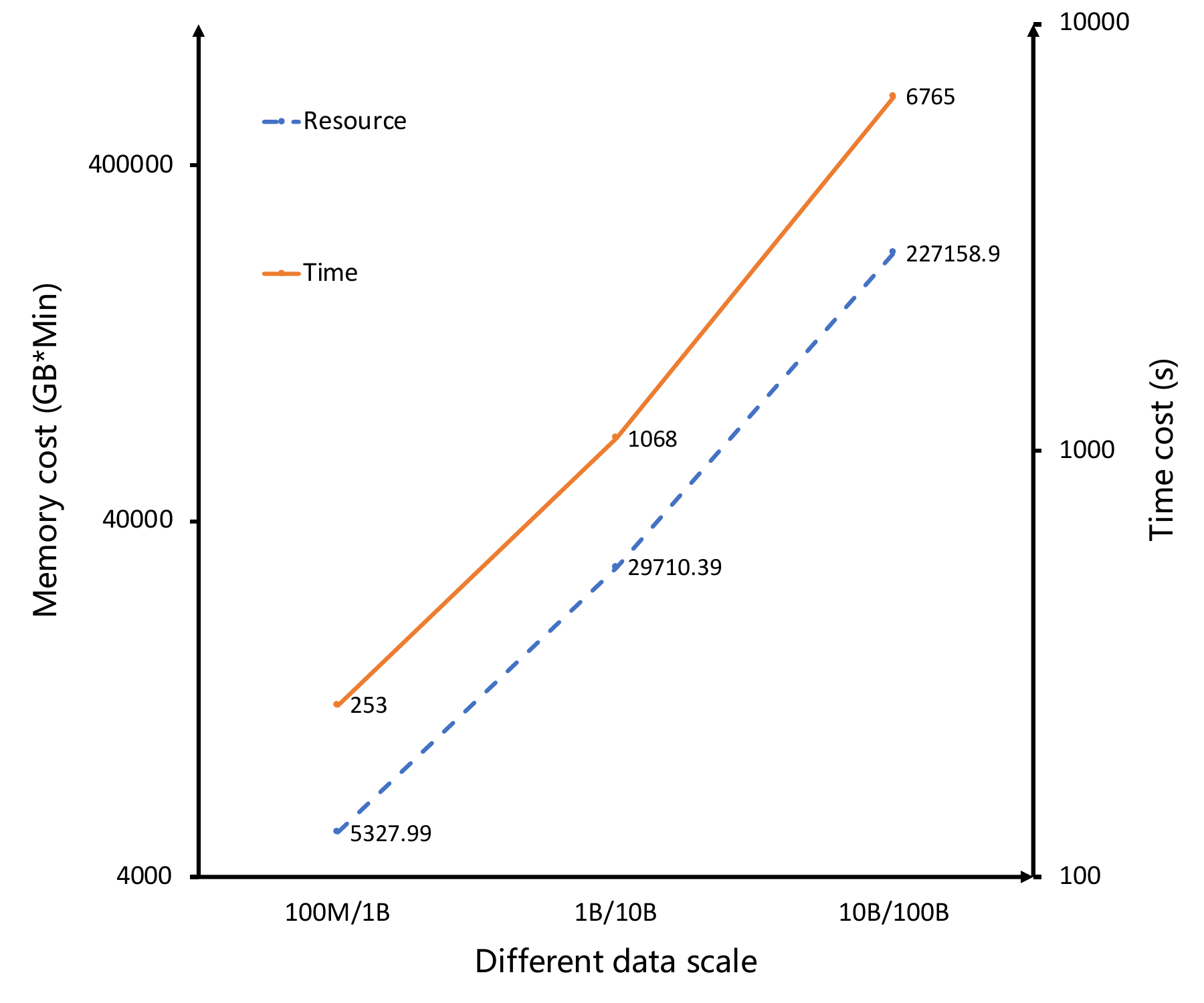}
\caption{Resource (cpu *min) and time cost (s) VS. DataScale}
\label{fig:scalability}
\end{minipage}
\hfill
\begin{minipage}{0.3\linewidth}
\centering
\includegraphics[width=\textwidth]{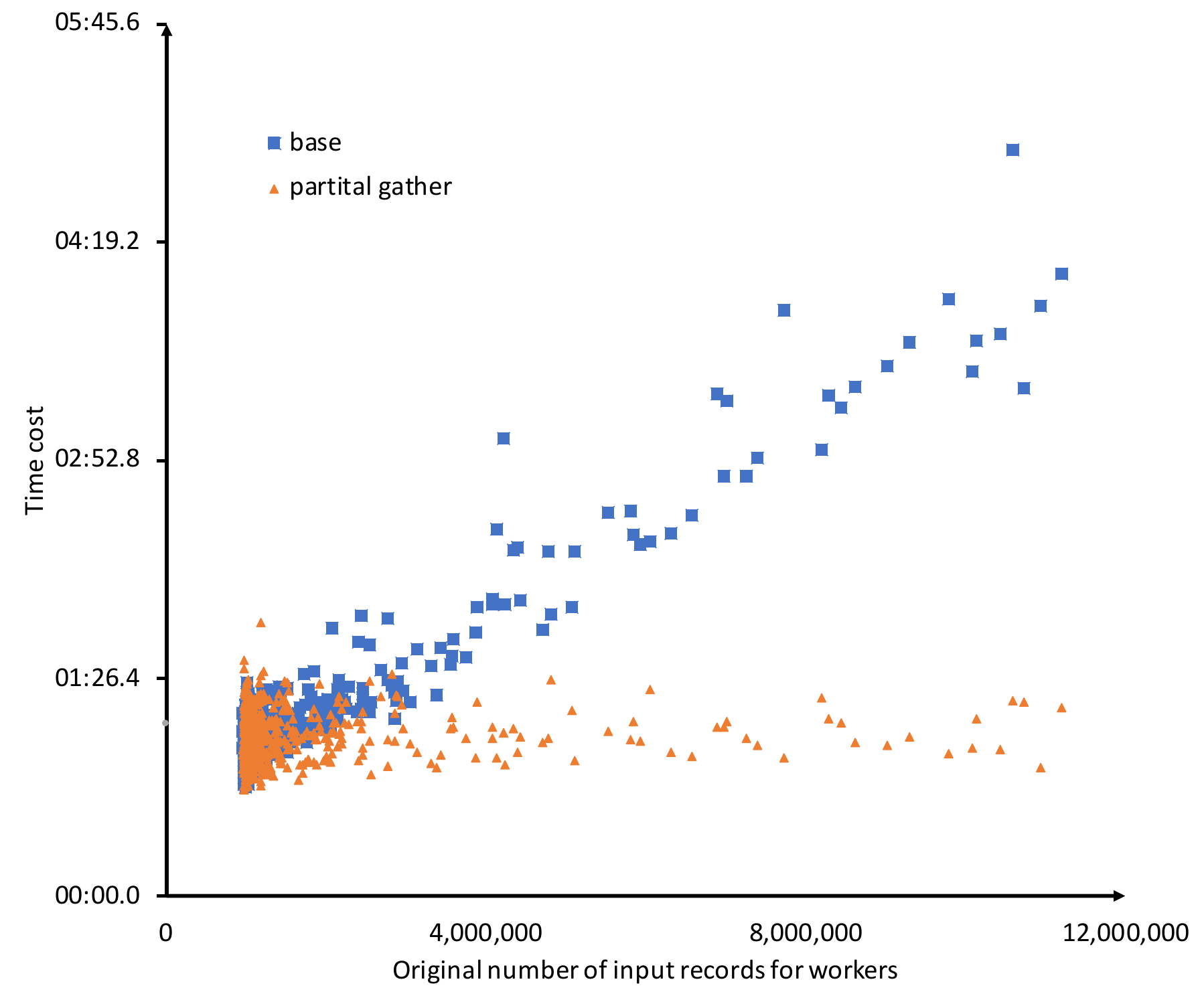}
 \caption{Time cost for different instances with partial gather strategy.}
\label{fig:paritial}
\end{minipage}
\hfill
\begin{minipage}{0.3\linewidth}
\centering
\includegraphics[width=\textwidth]{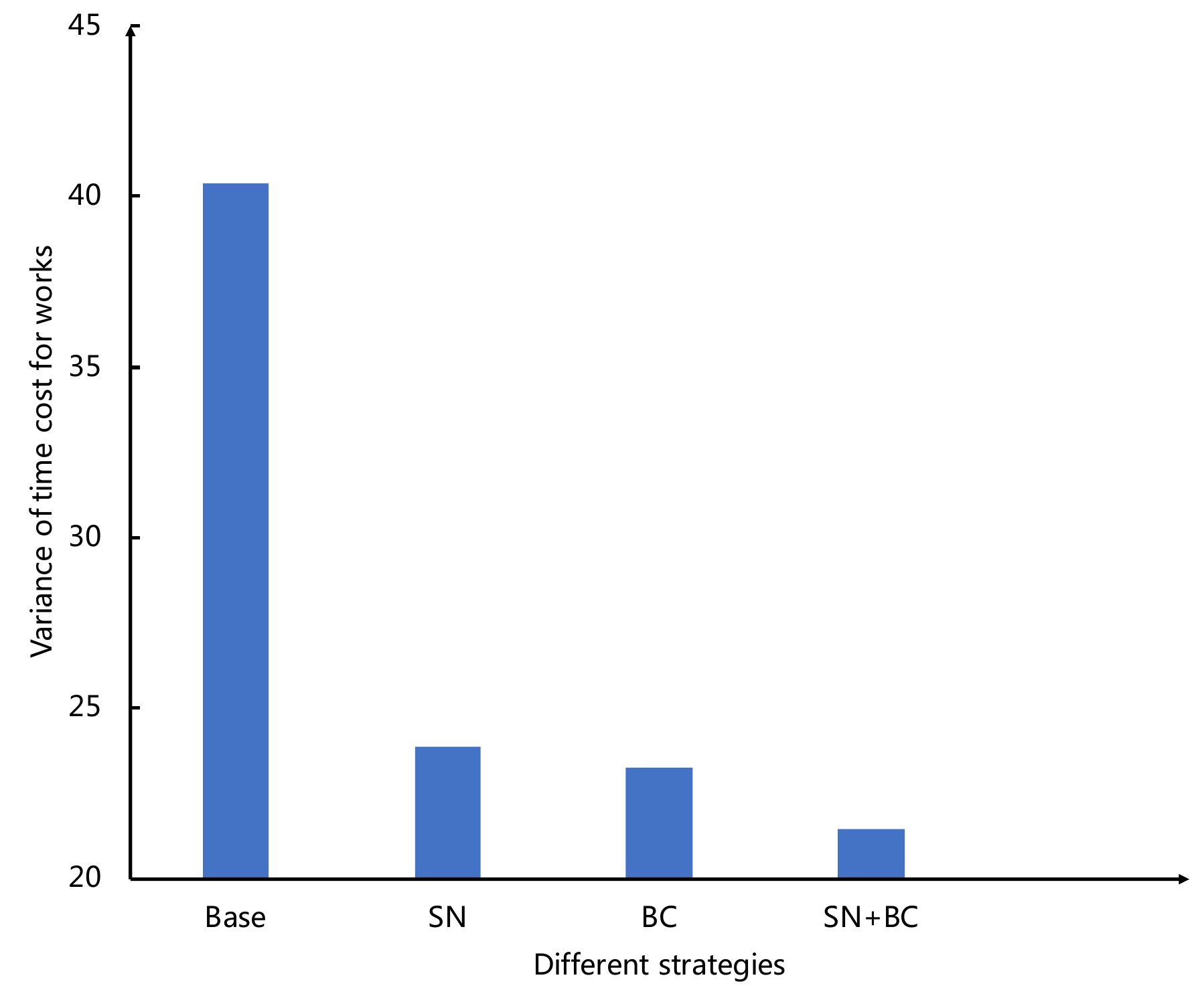}
 \caption{Variance of time cost with different strategies for nodes with large out-degree}
\label{fig:out_degree_strategies}
\end{minipage}
\end{figure*}

Figure~\ref{fig:scalability} demonstrates the experimental results. 
The y-axis on the left and right represent resource usage (cpu*min) and time cost (s), respectively, while the x-axis is the data scale.
Both the resource and the time-cost curves demonstrate a nearly linear relationship between them and the data scale, which proves the linear scalability of our system.
It is worth noting that our system could finish the inference task for all 10 billion nodes (with about 100 billion edges) within 2 hours (6765 seconds), which shows that our system could scale to extremely large graphs and finish inference tasks efficiently.




\begin{figure*}[htbp]
\centering
\begin{minipage}{0.3\linewidth}
\centering
\includegraphics[width=\textwidth]{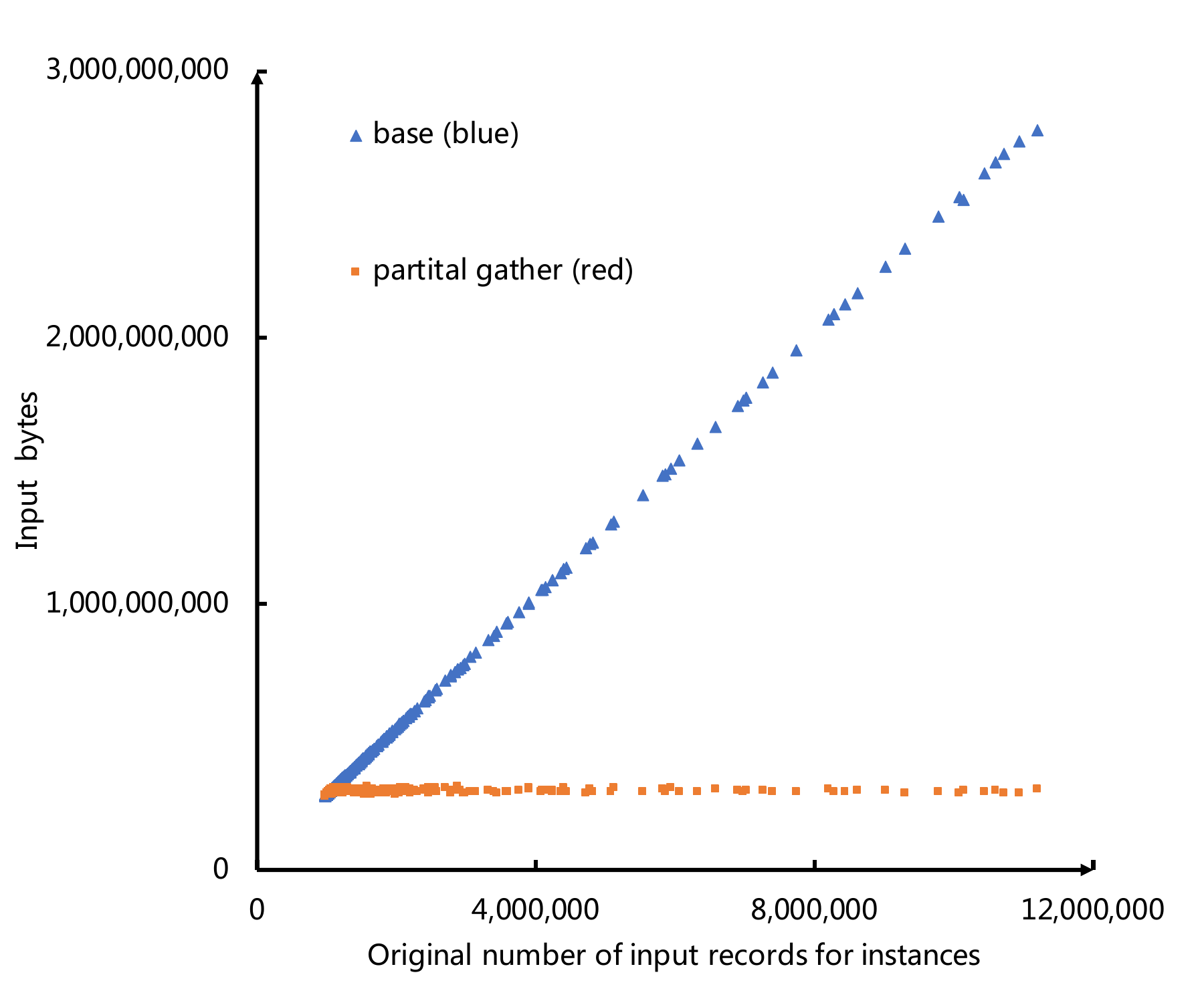}
\caption{IO cost for partial gather strategy}
\label{fig:in_degree_io_bytes}
\end{minipage}
\hfill
\begin{minipage}{0.3\linewidth}
\centering
\includegraphics[width=\textwidth]{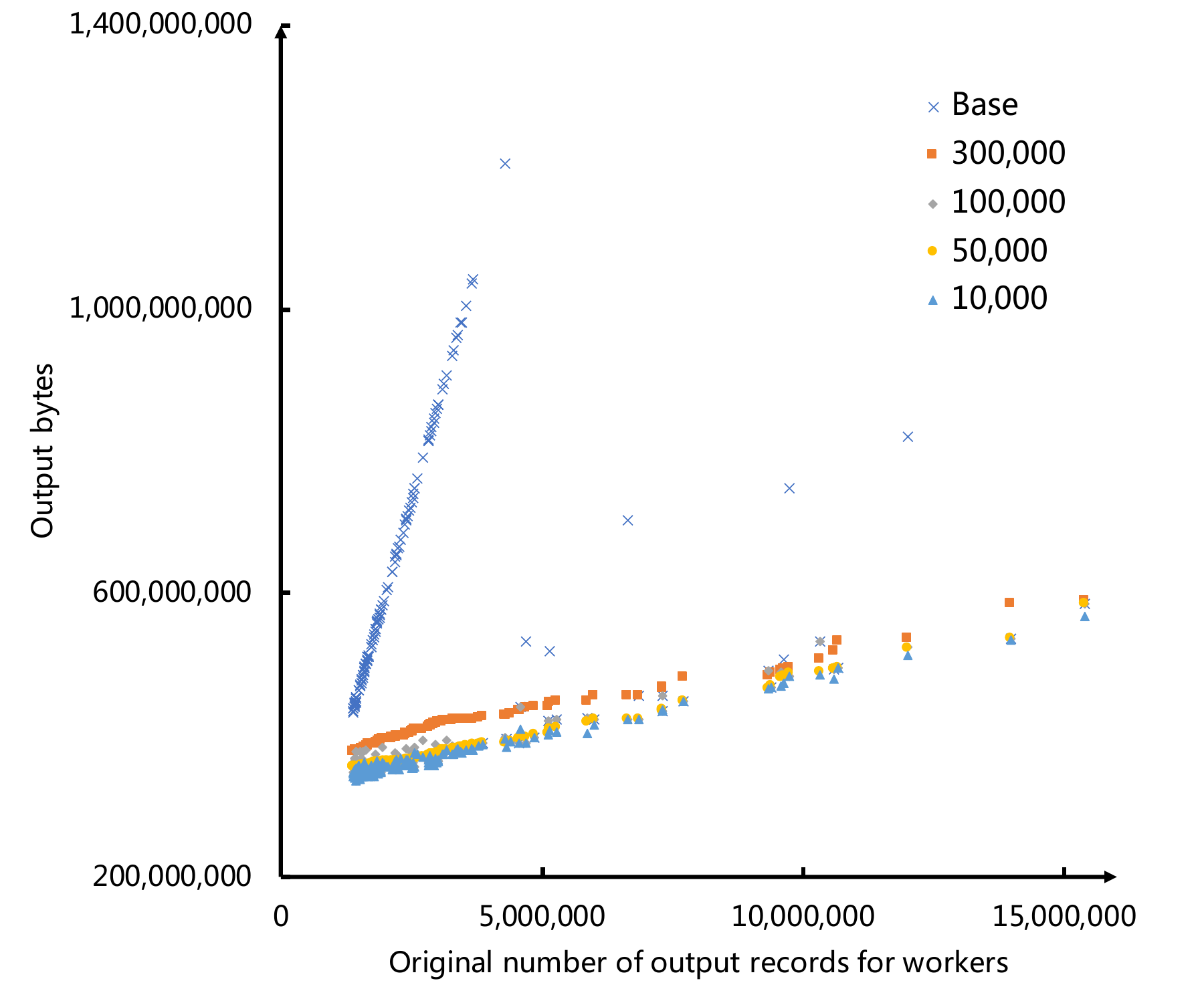}
 \caption{IO cost for broadcast strategy } 
\label{fig:out_degree_io_bytes}
\end{minipage}
\hfill
\begin{minipage}{0.3\linewidth}
\centering
\includegraphics[width=\textwidth]{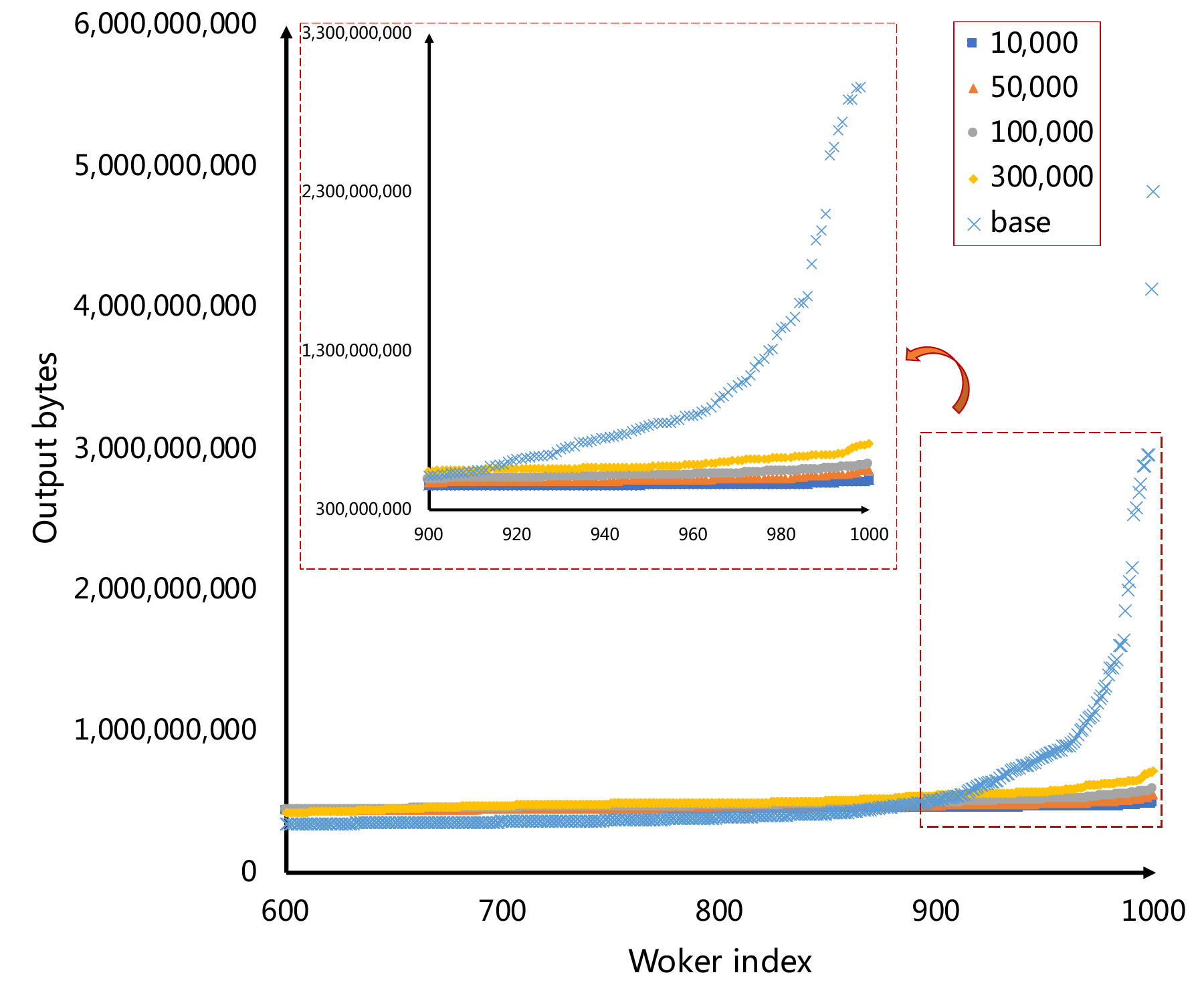}
 \caption{IO cost for shadow node strategy } 
\label{fig:out_degree_io_bytes_SN}
\end{minipage}
\end{figure*}

\subsubsection{Analysis for Optimization Strategies}
\label{sec_analyis_strategies}
For further analysis of our systems, we also design a set of experiments to verify the effectiveness of the optimization strategies for hub nodes proposed in Sec. \ref{sec_opt_stategies}.
We enable those strategies on the GraphSAGE model over Power-law datasets (about 100 million nodes, 1.4 billion edges) respectively for variable-controlling purposes.

Figure~\ref{fig:paritial} illustrates experimental results on the large in-degree problem and the effect of the \emph{partial-gather} strategy. 
The x-axis indicates the number of in-edges for nodes in one instance, while the y-axis denotes the instance latency, and each point represents an instance.
Results show that latency is positively associated with the number of in-edges.
It's reasonable since the time cost for receiving messages and the computation to aggregate those messages are related to the number of in-edges.
Therefore, nodes with large in-degree would lead the time cost of the related instance in the long tail and become the bottleneck of the whole inference task.
In contrast, the \emph{partial-gather} strategy helps release the stress caused by a large number of in-edges: the variance of time cost on different instances shrinks a lot, and points are close to the mean line.
The reason is from two aspects: firstly, with the \emph{partial-gather} strategy, messages would be partially aggregated on the sender side, which brings down the time cost in communication.
Secondly, the computation of in-edge messages is uniformly conducted in advance on different instances, which balances the computation load for hub nodes. 

We also make an analysis of the large out-degree problem, and the results are illustrated in Fig.~\ref{fig:out_degree_strategies}.
\emph{Base} denotes the experiment without optimization strategy, while \emph{SN}, \emph{BC}, and \emph{SN+BC} indicate experiments with \emph{shadow node} strategy, \emph{broadcast} strategy, and the combination of them respectively.
The y-axis represents the variance of time cost in all instances.
Compared with the baseline, both the \emph{shadow node} and \emph{broadcast} strategies are helpful in mitigating the problem caused by nodes with large out-edges, and the latter one is slightly better than the former one since \emph{shadow node} strategy would lead to some overhead by duplicating in-edges to average the communication over different mirror nodes.
Specially, for GraphSAGE, we could achieve better results by combining the two strategies since messages for different out-edges are the same.

Furthermore, we develop some experiments to evaluate our system's IO costs.
Figures~\ref{fig:in_degree_io_bytes},~\ref{fig:out_degree_io_bytes}, and~\ref{fig:out_degree_io_bytes_SN} show the results of strategies for large in- and out-degree problems, respectively.
Their y-axises show the input or output bytes of each instance.
For Fig.~\ref{fig:in_degree_io_bytes} and~\ref{fig:out_degree_io_bytes}, x-axises show the initial number of input or output records for different instances.
Specially, the x-axis for Fig.~\ref{fig:out_degree_io_bytes_SN} is worker index sorted according to the output bytes since the \emph{shadow node} strategy would re-arrange the placement of records. 

Figure~\ref{fig:in_degree_io_bytes} demonstrates how our system, using the \emph{partial-gather} approach, reduces the input IO cost to a constant level.
This approach is effective for all nodes regardless of degree and performs particularly well for nodes with large in-degree since there is no overhead involved.
The \emph{partial-gather} strategy reduces the total communication cost of all works by roughly 25\% while saving up to 73\% IO cost for the 10\% workers in the tail.
The reason is that each instance would only send one message for a particular node in the upcoming round since the messages have already been pre-aggregated.
Each node would only receive as many as instance-number messages in the following round.

Moreover, we also vary the threshold for those strategies to verify the heuristic formula in Sec.~\ref{sec_opt_stategies}.
The hyper-parameter $\lambda$ is set to 0.1 in our system, and the threshold to distinguish whether a node is a `hub' node then becomes 100,000 (1 billion edges, 1000 workers).
Compared with the baselines, both the \emph{broadcast} and \emph{shadow nodes} strategies in this setting could decrease the IO cost for ``hub'' nodes significantly.
For instance, the communication costs for the last 10\% workers are reduced by 42\% and 53\% when activating those two techniques, respectively.
Furthermore, the efficacy of those strategies would increase as the threshold decreases.
However, there is no significant difference for the threshold in the range of [10,000, 100,000], and the difference in IO cost is less than 5\%.
This is because a low threshold would enable more nodes to benefit from those techniques, but their overhead could not be disregarded.
For example, the memory cost is nearly doubled.
Therefore, although the threshold determined by the heuristic formula may not be the optimal one, it is reasonable to use it to quickly estimate the threshold, and in this setting, those strategies successfully reduce the IO cost for straggler workers.

In summary, the experimental results demonstrate that those strategies proposed in Sec.~\ref{sec_opt_stategies} help eliminate the stragglers caused by the power law, and as a result, our system achieves better load-balancing.

\section{Conclusion}
In this paper, we propose InferTurbo, to boost the inference tasks at industrial-scale graphs.
The design follows the GAS-like paradigm underlying the computation of GNNs, with which we could specify the data flow and computation flow of GNNs and hierarchically conduct the inference phase in full-graph manner.
In this way, our system could avoid the redundant computation problem.
We provide alternative backends of MapReduce and Pregel for different industrial applications, and both could scale to large graphs.
Furthermore, a set of strategies without dropping any information is used to solve problems caused by nodes with large in-degree or out-degree.
The experimental results demonstrate that our system achieves $30\times$ $\sim$ $50\times$ speedup compared with some state-of-the-art graph learning systems. Our system could finish the inference task over a graph with 10 billion nodes and 100 billion edges within 2 hours, which shows the superior efficiency and scalability of our system. 

\end{document}